\documentclass[conference]{IEEEtran}
\IEEEoverridecommandlockouts

\usepackage{cite}
\usepackage{amsmath,amssymb,amsfonts}
\usepackage{algorithmic}
\usepackage{graphicx}
\usepackage{textcomp}
\usepackage{xcolor}
\usepackage{url}
\usepackage{multirow}
\usepackage{subcaption}
\usepackage[linesnumbered,ruled,vlined]{algorithm2e}
\usepackage{tcolorbox}

\def\BibTeX{{\rm B\kern-.05em{\sc i\kern-.025em b}\kern-.08em
    T\kern-.1667em\lower.7ex\hbox{E}\kern-.125emX}}

\newcommand{\etal}{\textit{et al.} }
\newcommand{\ie}{\textit{i.e.}}
\newcommand{\eg}{\textit{e.g.}}
\newcommand{\ProjectName}[1]{{\small\textsc{DisPatch}}}

\newcommand{\revise}[2][black]{\textcolor{#1}{#2}}

\begin{document}

\title{\textsc{DisPatch}: Disarming Adversarial Patches in Object Detection with Diffusion Models}
\author{
\IEEEauthorblockN{Jin Ma, Mohammed Aldeen, Christopher Salas, Feng Luo, Mashrur Chowdhury, Mert Pes\'e, Long Cheng}
\IEEEauthorblockA{\textit{Clemson University}}
}


\maketitle

\begin{abstract}

Object detection is fundamental to various real-world applications, such as security monitoring and surveillance video analysis. Despite their advancements, state-of-the-art object detectors are still vulnerable to adversarial patch attacks, which can be easily applied to real-world objects to either conceal actual items or create non-existent ones, leading to severe consequences. 
In this work, we introduce \ProjectName{}, the first diffusion-based defense framework for object detection. Unlike previous works that aim to ``detect and remove'' adversarial patches, \ProjectName{} adopts a ``regenerate and rectify'' strategy, leveraging generative models to disarm attack effects while preserving the integrity of the input image. Specifically, we utilize the in-distribution generative power of diffusion models to regenerate the entire image, aligning it with benign data. A rectification process is then employed to identify and replace adversarial regions with their regenerated benign counterparts. \ProjectName{} is attack-agnostic and requires no prior knowledge of the existing patches.
Extensive experiments across multiple detectors demonstrate that \ProjectName{} consistently outperforms state-of-the-art defenses on both hiding attacks and creating attacks, achieving the best overall mAP@0.5 score of 89.3\% on hiding attacks, and lowering the attack success rate to 24.8\% on untargeted creating attacks. Moreover, it strikes the balance between effectiveness and efficiency, and maintains strong robustness against adaptive attacks, making it a practical and reliable defense method. 

\end{abstract}

\section{Introduction}\label{sec:intro}

Object detection stands as a cornerstone in computer vision, serving as the foundation for a wide range of applications such as surveillance systems~\cite{jain2024fusion}, robotics~\cite{achirei2023model}, autonomous driving~\cite{muhammad2020deep}, and augmented reality~\cite{ghasemi2022deep}. It allows artificial intelligence (AI) systems to recognize and locate objects in their surroundings. Modern machine learning (ML) object detection frameworks, such as YOLO~\cite{redmon2018yolov3}, Faster R-CNN~\cite{ren2016faster}, and DETR~\cite{carion2020end}, have achieved remarkable detection accuracy and efficiency, making them broadly applied to real-world AI systems. 
Despite their impressive performance, object detectors are vulnerable to adversarial attacks \cite{sharma2022adversarial,guesmi2023physical}, which exploit the inherent weaknesses of ML models to mislead them into producing incorrect output. 
Among various adversarial attack strategies, adversarial patch attacks have attracted significant attention from the security community, as they are easy to implement in the real world, and can cause severe consequences in object detection systems~\cite{kurakin2018adversarial,guesmi2023physical}. In adversarial patch attacks, the adversary creates patches that can be placed on real-world objects, such that when viewed by  object detectors, these patches trick the model into making incorrect predictions~\cite{eykholt2018robust, thys2019fooling, hu2021naturalistic, hu2022adversarial, hu2023physically}, leading to  \textit{hiding attacks} or \textit{creating attacks}. In hiding attacks~\cite{thys2019fooling, hu2021naturalistic, hu2022adversarial, hu2023physically, cao2023you}, the patch is applied to a victim object, making the object invisible to the detection system. For instance, in surveillance systems deployed in sensitive areas such as airports or government buildings, an adversarial patch affixed to clothing can prevent security cameras from recognizing prohibited individuals. 
In contrast, creating attacks~\cite{liu2022harnessing, yin2022adc, zhu2023tpatch} involve tricking the system into identifying a non-existent object. For example, an adversarial patch on a billboard could be used to make the surveillance system treat it as a suspect vehicle, leading to unnecessary investigations or divert attention from genuine security events. 

\begin{figure}
    \centering
    \includegraphics[width=0.95\linewidth]{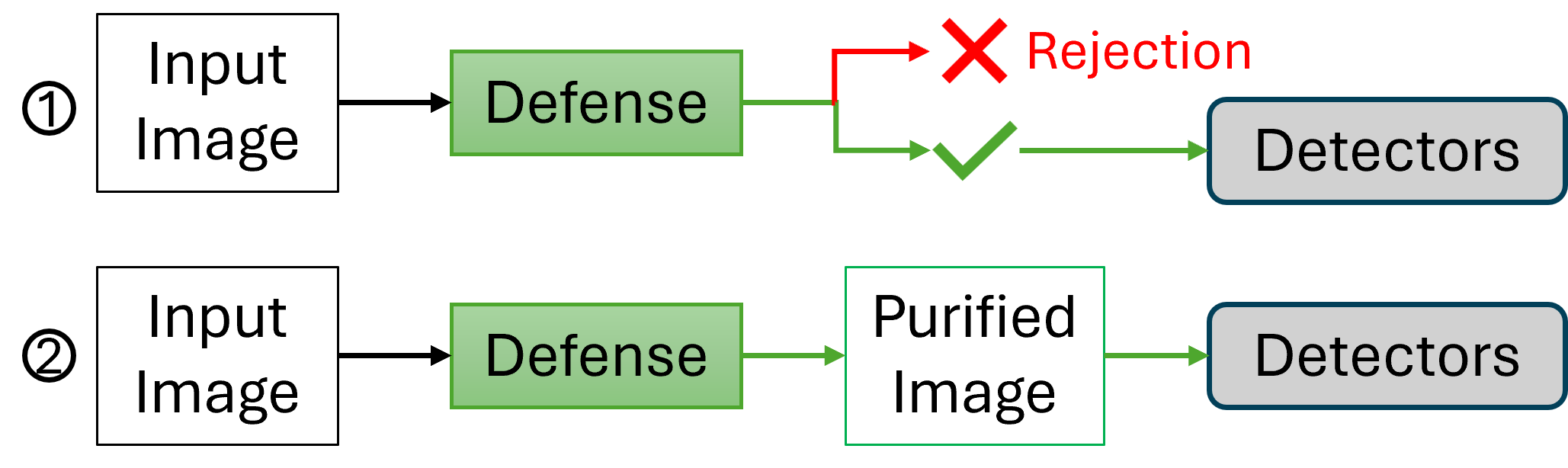}
    \caption{Two types of inference-level defenses: \textcircled{\small{1}} identifying and rejecting images with adversarial patches; \textcircled{\small{2}} purifying input images before object detection.}
    \label{fig:defense_approaches}
\end{figure}

In response to the threat of adversarial patch attacks, numerous defense mechanisms have been proposed, including training-level and inference-level defenses~\cite{noppel2024sok}. Training-level defenses focus on retraining the model on adversarial examples or designing more robust architectures. However, these methods require additional training time and may still fail to defend against unseen attacks. In contrast, inference-level defenses operate by pre-processing the input image before it is fed into the model, thereby eliminating the need for model retraining. Figure~\ref{fig:defense_approaches} illustrates two types of inference-level defenses: the first aims to identify and \textbf{reject} input images containing adversarial patches~\cite{xiang2021detectorguard,feng2024fight}; while the second \textbf{purifies} all input images before object detection~\cite{xiang2023objectseeker,lin2024don}. In this paper, we focus on the latter, as such methods do not interrupt the input data flow.

To purify input images that are potentially affected by adversarial patches, prior works typically adopt a ``detect and remove'' strategy: adversarial patches are first localized using specially designed methods or models, and then removed from the images~\cite{liu2022segment,tarchoun2023jedi,jing2024pad,lin2024don}. However, such approaches carry the risk of mistakenly removing benign regions, potentially degrading the detection performance. To address this limitation, we aim to develop a method that can instead ``disarm'' adversarial patches by transforming them into benign counterparts rather than removing them, thereby preserving the overall integrity of the input images. To this end, we use generative models to regenerate images as in-distribution data for object detectors, inspired by the observation that adversarial patches are generally non-robust or out-of-distribution (OOD) data for victim models~\cite{goodfellow2014explaining,cubuk2017intriguing}. Specifically, Diffusion Models (DMs), as powerful generative models, are trained to learn a generative process that captures the underlying training data distribution, enabling the data generation to be consistent with the clean training dataset~\cite{ho2020denoising,saharia2022image,nie2022diffusion}. 
By leveraging this in-distribution generative power of DMs, we propose \ProjectName{} to disarm adversarial patches and purify the input images.

\ProjectName{} encompasses a ``regenerate and rectify'' approach with two stages: \textit{regeneration stage} and \textit{rectification stage}. In the regeneration stage, \ProjectName{} employs an inpainting DM~\cite{suvorov2022resolution,rombach2022high} to regenerate the whole image. 
In the rectification stage, \ProjectName{} uses an adversarial detection algorithm to identify and replace potential adversarial regions with their benign counterparts, leaving clean regions untouched. 
\ProjectName{} does not rely on any prior knowledge of existing adversarial patches or the training of specific models. The only requirement is that the inpainting DM is trained on clean dataset, such that it can generate the in-distribution data for the detector. The main contributions of this paper are as follows:
\begin{enumerate}
    \item \textbf{An attack-agnostic and generalizable defense algorithm}: We propose a novel diffusion-based defense method \ProjectName{}\footnote{Code is available at \url{https://github.com/MaJinWakeUp/DisPatch}.}, which disarms adversarial patches for object detectors. \ProjectName{} leverages DMs to purify input images without requiring any prior knowledge of adversarial patches, making it a generalizable method for defending diverse adversarial patch attacks.

    \item \textbf{Unique ``regenerate and rectify'' approach}: Unlike most previous works that focus on ``detect and remove'' strategies, \ProjectName{} introduces a novel ``regenerate and rectify'' pipeline. This strategy disarms adversarial patches by neutralizing them into benign content through in-distribution generation, rather than explicitly removing them. Thus, it preserves the integrity of input images.  
    %

    \item \textbf{Enhanced defense performance and robustness}: Extensive experiments demonstrate that \ProjectName{} achieves state-of-the-art performance across different detectors and attacks. It attains the highest overall mAP@0.5 score against \textit{hiding attacks}, and the greatest reduction of attack success rates against \textit{creating attacks} over other baselines. Moreover, \ProjectName{} shows strong robustness against adaptive attacks, indicating its resilience under stronger and more targeted threat scenarios.
\end{enumerate}

\begin{figure*}[ht]
    \centering
    \includegraphics[width=\textwidth]{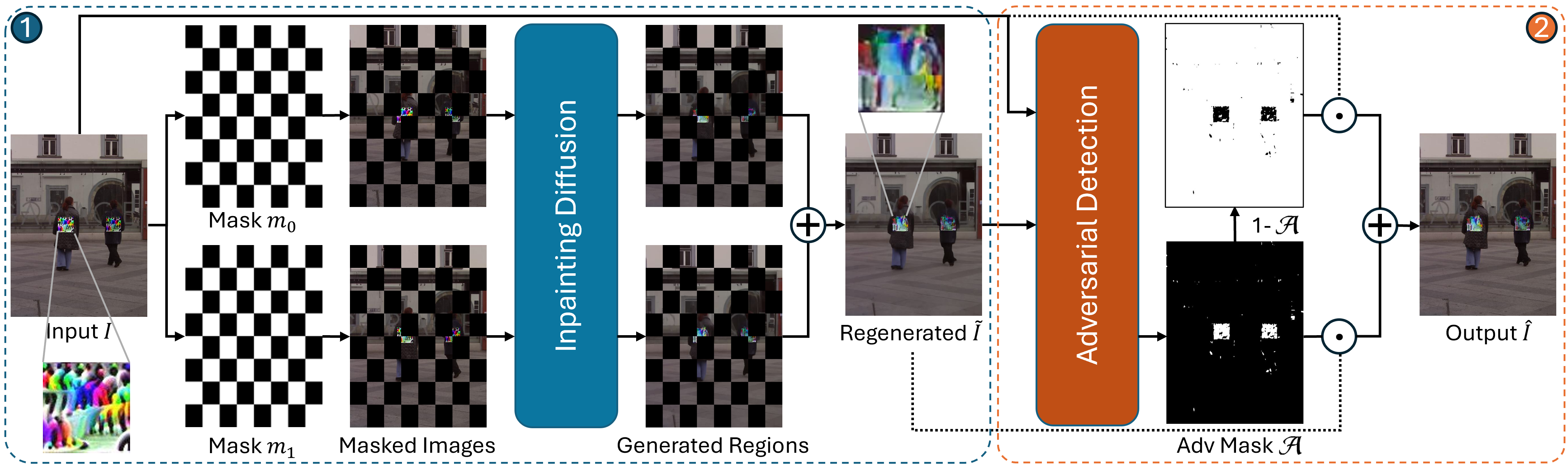}
    \caption{The \ProjectName{} pipeline consists of two key stages: \textcircled{\small{1}} The regeneration stage, which generates a fully reconstructed image to disarm adversarial patches. \textcircled{\small{2}} The rectification stage, which replaces adversarial regions in the input image with generated benign counterparts, according to the adversarial mask predicted by our adversarial detection algorithm.}
    \label{fig:pipeline}
\end{figure*}

\section{Related Work}\label{sec:related}
\subsection{Adversarial Patch Attacks}
Adversarial patch attacks pose a significant threat to object detection systems. Current patch attacks can be categorized into \textit{hiding attacks} and \textit{creating attacks}. For hiding attacks, Thys \etal introduced printable adversarial patches that are capable of concealing individuals from object detectors~\cite{thys2019fooling}. These patches are easily noticeable to human observers, limiting their stealthiness.
To address this limitation, researchers have explored various strategies to enhance the naturalness of adversarial patches. For instance, Huang \etal introduced patches with camouflage patterns by incorporating constraints during optimization~\cite{huang2020universal}, and Hu \etal deployed Generative Adversarial Networks (GANs) to generate patches that feature common imagery like animals~\cite{hu2021naturalistic}. Other approaches focus on utilizing real-world objects with specifically designed textures to evade detection. For example, Hu \etal developed adversarial textures that can be applied to clothing in arbitrary shapes~\cite{hu2022adversarial}, Zhu \etal also proposed a novel adversarial camouflage texture using 3D modeling techniques to achieve a high degree of naturalness~\cite{hu2023physically}. 
When crafting creating attacks, a commonly used approach is the Expectation over Transformation (EOT) technique, which optimizes a patch with random transformations~\cite{zhao2019seeing,braunegg2020apricot}.
These advanced and diverse attacks expose serious weaknesses in object detectors, highlighting the urgent need for more adaptable defenses.

\subsection{Inference-level Defenses}
As shown in Figure~\ref{fig:defense_approaches}, the first type of inference-level defenses focuses on identifying and rejecting images with adversarial patches. For instance, Xiang \etal developed an objectness predictor and objectness explainer to detect unexplained objectness caused by the adversarial patches, and to trigger an attack alert when such anomalies occur~\cite{xiang2021detectorguard}. Feng \etal trained two defensive patches, canary and woodpecker, to probe potential adversarial patches~\cite{feng2024fight}. These approaches only raise an alert when adversarial patches exist, but cannot help detectors make correct predictions under these attacks, limiting their practical applications. In contrast, the second type aims to purify input images without interrupting the input data flow. 
For example, Xiang \etal introduced a patch-agnostic masking strategy that removes adversarial patches without prior knowledge of their shape, size, or location~\cite{xiang2023objectseeker}. While effective within specific threat models, this approach is limited to scenarios involving only a single adversarial patch per image. Liu \etal developed a patch segmentation model to identify adversarial patches, followed by a shape-completion step to refine detection and remove the patches from images~\cite{liu2022segment}. Tarchoun \etal identified adversarial patches by detecting entropy peaks in the image and used an autoencoder to inpaint affected areas~\cite{tarchoun2023jedi}. \revise{Chen \etal developed a two-stage defense method to first detect adversarial patches with hold-out images, and then mitigate adversarial patches with random masked GAN generation\cite{chen2023jujutsu}}. Jing \etal utilized semantic independence and spatial heterogeneity, supported by the Segment-Anything-Model (SAM)~\cite{kirillov2023segment}, to effectively localize patches~\cite{jing2024pad}. Lin \etal proposed a block-based method in which an autoencoder was used to detect adversarial regions by rejecting out-of-distribution blocks~\cite{lin2024don}. Our \ProjectName{} differs from previous works in two aspects. First, instead of following the conventional ``detect and remove\revise{/mitigate}'' strategy, we adopt a ``regenerate and rectify'' strategy to preserve the integrity of input images. Second, \ProjectName{} operates directly on off-the-shelf DMs and does not require any training of customized models.

\subsection{Diffusion-based Defenses}
In this work, we address adversarial patch defense by leveraging the observation that adversarially perturbed data diverges from the distribution of clean data, which has been explored in previous adversarial detection research~\cite{nie2022diffusion,lin2024don}. \revise{DMs are well-suited for generating \textit{in-distribution} data because they are trained as likelihood-based generative models to reverse a fixed forward noising process that maps clean samples to a simple prior (\eg, Gaussian noise), learning this reverse denoising dynamics yields a sampling procedure that transports noise back toward high-density regions of the \textit{training data distribution}~\cite{sohl2015deep,ho2020denoising,songs2020core}. This training is tightly connected to likelihood-based density modeling~\cite{ho2020denoising,song2021maximum}, and empirically DMs achieve high-fidelity generation with strong distribution coverage~\cite{nichol2021improved,dhariwal2021diffusion}. This in-distribution generation property can be utilized to mitigate adversarial data. For example, Nie \etal demonstrated that adversarial perturbations could be washed out by the forward process in DMs, supported by both theoretical analysis and empirical results, and accordingly proposed DiffPure to purify adversarially perturbed images~\cite{nie2022diffusion}. Liu \etal proposed LDM, which leverages inpainting DMs to detect OOD data by comparing the inpainted image with the original image, and then the flagged OOD images are rejected from further processing~\cite{liu2023unsupervised}.}

\revise{\ProjectName{} similarly leverages diffusion models to process input images, but differs from DiffPure~\cite{nie2022diffusion} in two key ways: (i) it introduces an additional rectification stage beyond diffusion-based regeneration stage, and (ii) it is tailored to object detection task, whereas DiffPure is designed for image classification task. \ProjectName{} also differs from LDM~\cite{liu2023unsupervised} in that it preserves usability of the processed images for downstream prediction, whereas LDM focuses on OOD detection and rejects flagged samples rather than repairing them.}

\section{Threat Model and System Design}


\subsection{Threat Model}
\textbf{Attacker.} \revise{The objective of an adversary is} to deceive victim object detectors by introducing adversarial patches into input images. These patches can either be digitally added to the image or physically applied to real-world objects, impacting the images captured by cameras. The adversary has the flexibility to modify the size, shape, and location of the patches to achieve one of two objectives: \textit{hiding attacks}, where existing objects are concealed; and \textit{creating attacks}, where non-existent objects are fabricated. 
We assume a strong adversary with full knowledge of the object detector’s architecture and weights, enabling a white-box attack. In addition, the adversary is aware of the defense method, \ProjectName{}, and can design \textit{adaptive attacks} accordingly. However, the adversary is not permitted to tamper with or modify the object detector or the defense model itself.

\revise{\textbf{Defender.} The defender aims to mitigate adversarial patches through input pre-processing. Given an image, the defender regenerates it with a diffusion model to project the input toward the benign data distribution. We assume the defender has access to the benign training distribution of the victim detector and trains the diffusion model on the same data. This reflects practical deployments where the defense is implemented by, or requested from, the detector developer who typically controls the training data.}

\subsection{System Design}

Figure \ref{fig:pipeline} illustrates the pipeline of \ProjectName{}, which comprises two key stages: \textcircled{\small{1}} regeneration stage, \textcircled{\small{2}} rectification stage. In the regeneration stage, the entire input image is regenerated through an inpainting DM, using two unique inpainting masks. The goal in this stage is to produce a reconstructed image containing only in-distribution data, effectively removing any adversarial artifacts. However, the regenerated image cannot be directly used for detection, since the benign objects may be affected during the regeneration process. \textit{We aim to only replace the possible adversarial regions with the regenerated counterparts, while keeping the others in their original form.} In the rectification stage, the regenerated image is compared with the original input image to identify potential adversarial regions by using an effective adversarial detection algorithm. After obtaining the adversarial mask $\mathcal{A}$, the suspected regions in the original input are replaced with the corresponding content from the regenerated image. This two-stage process ensures that the final output is both free of adversarial influences and retains the integrity of clean regions.


\section{The Proposed Method}
\subsection{Regeneration Stage}
To regenerate the entire image efficiently and ensure it contains only benign in-distribution pixels, we use an off-the-shelf inpainting DM to perform this task. The inpainting DM takes the original image and a mask as input, and then generates new content in the masked areas based on contextual information from the unmasked regions. Since the inpainting process relies on the unmasked regions as context, only a partial image can be regenerated in each forward pass. This leads to a key challenge: \textit{``How can we regenerate the entire image with the minimum number of forward passes while preserving its original content?''}


\textbf{Efficient regeneration strategy.} We use two complementary binary masks: the checkerboard mask $m_0$ and its inverse checkerboard mask $m_1$, as shown in Figure \ref{fig:pipeline}. Inspired by prior work~\cite{liu2023unsupervised}, these masks are designed such that the entire image can be regenerated with only two inpainting passes. Each mask consists of $N\times N$ grids, where a value of 1 (white) indicates a region to be regenerated, and 0 (black) denotes a preserved region. We enforce $m_0 \oplus m_1 = J$, where $\oplus$ denotes element-wise summation and $J$ is an all-ones matrix, ensuring that the two masks together cover all regions in the image. 



\textbf{Regeneration process.} Unlike prior work~\cite{liu2023unsupervised} that uses a pool of such masks to generate multiple images for comparison with the original image, here we use only two masks to obtain a fully regenerated image. Given an input image $I$, and two masks $m_0$ and $m_1$, we process the image in parallel through the inpainting DM $\mathcal{D}$ to obtain two inpainted images $\tilde{I}_0$ and~$\tilde{I}_1$:
\begin{equation}
    \tilde{I}_0=\mathcal{D}(I;m_0), \quad\tilde{I}_1=\mathcal{D}(I;m_1)
\end{equation}
The final regenerated image $\tilde{I}$ is then formed by combining the inpainted regions from $\tilde{I}_0$ and $\tilde{I}_1$ according to their masks:
\begin{equation}
    \tilde{I} = \tilde{I}_0\odot m_0 \oplus \tilde{I}_1\odot m_1
\end{equation}
where $\odot$ denotes element-wise multiplication.
By using this efficient masking strategy, we regenerate the entire image with only two forward passes, minimizing computational overhead. Since every region of the image is processed in this stage, adversarial patches are effectively transformed into benign and in-distribution data regardless of their location or shape.


\subsection{Rectification Stage}\label{Rectification}
In this stage, we aim to find the potential adversarial regions and replace them with the corresponding regenerated content, while keeping the benign regions untouched. Given the properties of the diffusion network, the regenerated image $\tilde{I}$ is inherently an in-distribution image. As a result, the regenerated regions of adversarial patches tend to exhibit a different distribution from their original adversarial form. As illustrated in the two zoom-in views in Figure~\ref{fig:pipeline}, the adversarial patch in $\tilde{I}$ differs significantly from the corresponding patch in $I$. Conversely, the regenerated benign regions in $\tilde{I}$ should closely align with the distribution of original benign regions in $I$. Leveraging this characteristic, we propose an effective adversarial detection method.

\textbf{Adversarial detection.}
As shown in Algorithm~\ref{alg:adversarial_detection}, we first normalize both the original image $I \in \mathbb{R}^{3 \times W \times H}$ and the regenerated image $\tilde{I} \in \mathbb{R}^{3 \times W \times H}$ to the range of $[0,1]$, where $3$ is the RGB channels, $W$ and $H$ are the sizes of the image. Then, we compute the pixel-wise L2 distance matrix $D \in \mathbb{R}^{W \times H}$, where each element in $D$ is computed as:
\begin{equation}
    D[i, j] = \parallel I[i, j] - \tilde{I}[i, j] \parallel_2, \quad 1 \leq i \leq W, 1 \leq j \leq H
\end{equation}
$I[i, j]$ and $\tilde{I}[i, j]$ represent the RGB vectors at pixel $(i, j)$. Thus, $D$ reflects the pixel-wise difference between the regenerated image and the original image. To mitigate the noise introduced by the stochastic nature of image generation, we apply \revise{Gaussian Smoothing}~\cite{gonzalez2009digital} to the distance matrix $D$, reducing local fluctuations and smoothing adjacent regions.


\begin{algorithm}[t]
\caption{Adversarial Detection Algorithm}
\label{alg:adversarial_detection}
\SetAlgoLined
\KwIn{$I \in \mathbb{R}^{3 \times W \times H}$, $\tilde{I} \in \mathbb{R}^{3 \times W \times H}$}
\KwOut{Adversarial mask $\mathcal{A} \in \mathbb{R}^{W \times H}$}
Normalize $I$ and $\tilde{I}$ to $[0,1]$\;
Compute distance matrix $D[i, j] = \parallel I[i, j] - \tilde{I}[i, j] \parallel_2, \; \forall i, j$\;
\revise{Apply Gaussian Smoothing to $D$ to reduce noise}\;
Flatten $D$ into vector $\mathbf{d} \in \mathbb{R}^{W \cdot H}$\;
Apply KMeans clustering on $\mathbf{d}$ to obtain clusters $\phi_0$ (with smaller centroid) and $\phi_1$ (with larger centroid)\;
\For{$i = 1$ \textbf{to} $W$}{
    \For{$j = 1$ \textbf{to} $H$}{
        $\mathcal{A}[i, j] \gets 1$ if $D[i, j] \in \phi_1$, else $\mathcal{A}[i, j] \gets 0$\;
    }
}
\Return $\mathcal{A}$
\end{algorithm}




\revise{Given the pixel-wise difference matrix $D$ between the regenerated image and original image. Intuitively, pixels covered by an adversarial patch tend to change more after regeneration, thus with high difference values, while benign pixels change less, thus exhibit low values. We then leverage this property to localize potential adversarial patches. Specifically, we flatten $D$ into a vector $\mathbf{d} \in \mathbb{R}^{W \cdot H}$ and apply KMeans clustering~\cite{macqueen1967some} with $k=2$, which automatically splits all pixels into two clusters $\phi_0$ and $\phi_1$ based on their difference values. We denote $\phi_0$ as the cluster with the smaller centroid, corresponding to pixels with low difference values that we treat as benign. Conversely, $\phi_1$ is the cluster with the larger centroid, corresponding to pixels with high difference values that we treat as adversarial.} Then we construct the adversarial mask $\mathcal{A}$ by marking all pixels belonging to $\phi_1$ as adversarial:
\begin{equation}
    \mathcal{A}[i, j] = 
    \begin{cases}
    0 & : D[i, j] \in \phi_0 \\
    1 & : D[i, j] \in \phi_1
    \end{cases}
    ,\quad 1 \leq i \leq W, 1 \leq j \leq H
\end{equation}

\textbf{Final output.} According to the adversarial mask $\mathcal{A}$ from Algorithm~\ref{alg:adversarial_detection}, we combine $I$ and $\tilde{I}$ to obtain the final output:
\begin{equation}\label{equ:rectify}
    \hat{I} = \mathcal{A} \odot \tilde{I} \oplus (1-\mathcal{A})\odot I
\end{equation}
In this step, those potential adversarial pixels identified by $\mathcal{A}$ are replaced with their generated counterparts in $\tilde{I}$, while the remaining pixels retain their original values in $I$. 

This approach enables the proposed algorithm to efficiently identify adversarial regions within an image by leveraging the differences between the original input image and the regenerated image. The clustering step automatically segments pixels into adversarial and benign groups, eliminating the need for prior thresholding or manual intervention. It is worth mentioning that even if some benign regions are mistakenly identified as adversarial, the rectification process does not degrade their quality. This resilience comes from the design of \ProjectName{}, which replaces suspect regions rather than removing them entirely. As a result, the overall visual fidelity of benign regions is preserved after processing.

\subsection{Diffusion Networks}~\label{sec:method_training}
To reduce the time complexity of the regeneration stage in \ProjectName{}, we adopt the Latent Diffusion Model (LDM)~\cite{rombach2022high} as the backbone for inpainting. LDM operates in a latent space, where autoencoders process a compressed representation of the image instead of working directly in the image space. This design significantly improves the efficiency of the generation process while maintaining high-quality reconstructions, making LDM an ideal choice for \ProjectName{}. \revise{To balance generation quality and computational efficiency, we resize all images to $512\times512$ for the inpainting process, which is a standard resolution for DMs. Using a lower resolution such as $256\times256$ can further reduce runtime, but it is substantially smaller than typical object detector input sizes (\eg, YOLOv3~\cite{redmon2018yolov3} uses $416\times416$), which may degrade downstream object detection performance. In contrast, a higher resolution such as $1024\times1024$ can improve visual fidelity but incurs larger computational overhead.}

\section{Design of Experiments}\label{sec:exp}
To evaluate the defense performance of \ProjectName{}, we organize our experiments into three parts: defending hiding attacks, defending creating attacks, and real-world validation. For each part, we describe the datasets, victim object detectors, attack methods, and evaluation metrics used. For comparison, we include several state-of-the-art defenses, including a certifiable defense method ObjectSeeker (OS)~\cite{xiang2023objectseeker}, and four empirical defense methods SAC~\cite{liu2022segment}, Jedi~\cite{tarchoun2023jedi}, PAD~\cite{jing2024pad}, and NutNet~\cite{lin2024don}.
Given the inherent randomness of diffusion models, all results for the proposed \ProjectName{} are reported as the average of three runs. 
The default number of grids is set to $N=32$, and the default sampling steps of LDM is set to $s=5$ in our experiments (ablation studies in Sec.~\ref{sec:eval_ablation}). All experiments were performed on a server with an NVIDIA A100 GPU.  


\subsection{Defending Hiding Attacks}
\textbf{Dataset.} We use INRIA-Person~\cite{dalal2005histograms} dataset for evaluation in this part, which includes 614 images for training and 288 images for testing, where each image contains at least one person. This dataset has been utilized in previous studies for developing various attack methods on person detectors~\cite{thys2019fooling,hu2021naturalistic,hu2023physically}, and for assessing defense methods against these attacks~\cite{tarchoun2023jedi,jing2024pad,lin2024don}.

\textbf{Victim object detectors.} We select three representative object detectors as the victim detectors: YOLOv3~\cite{redmon2018yolov3}, Faster RCNN~\cite{ren2016faster}, and DETR~\cite{carion2020end}. YOLOv3 and Faster RCNN are two CNN-based object detectors, with YOLOv3 representing single-stage detectors and Faster RCNN representing two-stage detectors. DETR is a Transformer-based object detector. In our experiment, we implement these detectors using the MMDetection toolbox~\cite{mmdetection} and utilize pretrained models on the MS-COCO~\cite{lin2014microsoft} dataset. 
\revise{Notably, the inpainting LDM was also trained on MS-COCO dataset, consistent with viction object detectors. In this way, we can ensure the generated data aligns with the training data distribution for detectors.}

\textbf{Attack methods.} Person-hiding attacks have been extensively studied due to their diverse real-world manifestations and significant security implications. In our experiments, we select three representative attack methods: adversarial patch attack (AdvPatch)~\cite{thys2019fooling}, naturalistic patch attack (NatPatch)~\cite{hu2021naturalistic}, and adversarial texture attack (AdvTexture)~\cite{hu2022adversarial}. These methods vary in form and attack capability.
To ensure consistency, we retrain these attack methods on the training data of INRIA-Person, targeting the three victim object detectors \revise{for generating adversarial pathces}. Then we apply the trained patches to the testing data for evaluation. The adversarial patch is applied to every person within test images. For AdvPatch and NatPatch, we set the patch size to 0.2, following the settings provided in their open-source code~\cite{advpatch}. This scale ratio represents the patch height relative to the diagonal length of the victim object bounding box. For AdvTexture, we adjust the scale ratio to 0.25, as it simulates human clothing, which covers more regions of the victim person. We will show that \ProjectName{} is robust to changes of patch size in Sec.~\ref{sec:eval_ablation}. 

\textbf{Evaluation metrics.} We report two metrics for this task. The first is mean average precision (mAP) at an intersection-over-union (IoU) threshold of 0.5, a standard metric for assessing object detection performance~\cite{jing2024pad,lin2024don}. The second is average recall (AR), which measures the average fraction of ground-truth objects successfully detected across different IoU thresholds, reflecting the detector’s ability to recover true positives. For both mAP@0.5 and AR, higher values indicate better defense performance.

\subsection{Defending Creating Attacks}
\textbf{Dataset.} We use an existing benchmark APRICOT~\cite{braunegg2020apricot} to evaluate the defense performance of \ProjectName{} under creating attacks. APRICOT provides a collection of \revise{real-world} photos with physical adversarial patches in a variety of scenarios, where the patches \revise{were} trained to create non-existent objects. The 873 test images in the benchmark were directly used for evaluation, where each image contains one adversarial patch with different shapes. We exclude the ObjectSeeker~\cite{xiang2023objectseeker} from the comparison methods here since it is specifically designed to defend against hiding attacks. 

\textbf{Victim object detectors.} The adversarial patches in APRICOT were trained to attack three different object detectors, Faster RCNN~\cite{ren2016faster}, SSD~\cite{liu2016ssd}, and RetinaNet~\cite{ross2017focal}. \revise{Similarly, we implement these detectors with MMDetection toolbox and use pretrained models on MS-COCO dataset.} 
In our experiments, we focus on white-box attacks, which means the attacker knows the details of victim object detector. The goal of the attacker is to launch either untargeted or targeted attacks. Untargeted attacks make the detector predict a non-existent object based on the adversarial patch, regardless of the predicted object class, while targeted attacks make the detector predict a specific target class based on the adversarial patch. 


\textbf{Evaluation metrics.} We evaluate defense performance using mAP@0.5 and attack success rate (ASR). For mAP@0.5, we treat the adversarial patch as the ground-truth object. In this setting, a lower mAP@0.5 indicates better defense, since the detector is less likely to recognize the patch as a valid object. For ASR, we follow APRICOT~\cite{braunegg2020apricot}, where an attack is defined as successful if ``at least one bounding box is wrongly predicted by the victim object detector, which has confidence over 0.3 and IoU of 0.1 with the adversarial patch''. ASR thus measures the fraction of adversarial examples that successfully fool the detector, and a lower ASR indicates stronger defense.



\begin{table*}[!t]
    \centering
    \caption{The performance of \ProjectName{} compared with other defense methods against hiding attacks on INRIA-Person benchmark. $^\dag$ indicates the certifiable defense method, while the others are empirical defense methods. OS is not applicable to YOLOv3, so we do not report the results in this table. Here, mAP refers to mAP@0.5.}
    \label{tab:inria}
    \begin{tabular}{|c|c|c|c|c|c|c|c|c|}
        \hline
        \multirow{2}{*}{Object Detector} & \multirow{2}{*}{Attack} & No Defense & \multicolumn{6}{c|}{With Defense} \\
        \cline{3-9}
          &  & Vanilla & OS$^\dag$~\cite{xiang2023objectseeker} & SAC~\cite{liu2022segment} & Jedi~\cite{tarchoun2023jedi} & NutNet~\cite{lin2024don} & PAD~\cite{jing2024pad} & \ProjectName{} \\ 
          &  & mAP \hfill AR & mAP \hfill AR & mAP \hfill AR & mAP \hfill AR & mAP \hfill AR & mAP \hfill AR & mAP \hfill AR \\ 
        \hline
        \multirow{4}{*}{YOLOv3} 
          & Clean                               & 0.910 \hfill \textbf{0.695} & - \qquad - & 0.900 \hfill 0.672 & \textbf{0.935} \hfill 0.668 & 0.919 \hfill 0.687 & 0.909 \hfill \underline{0.688} & \underline{0.934} \hfill 0.680  \\ 
          \cline{2-9}
          & AdvPatch~\cite{thys2019fooling}     & 0.498 \hfill 0.501 & - \qquad - & 0.579 \hfill 0.514 & 0.750 \hfill 0.558 & 0.808 \hfill 0.625 & \underline{0.811} \hfill \underline{0.626} & \textbf{0.903} \hfill \textbf{0.649} \\
          & NatPatch~\cite{hu2021naturalistic}  & 0.707 \hfill 0.593 & - \qquad - & 0.687 \hfill 0.562 & 0.760 \hfill 0.555 & 0.746 \hfill 0.597 & \underline{0.819} \hfill \textbf{0.633} & \textbf{0.840} \hfill \underline{0.605} \\
          & AdvTexture~\cite{hu2022adversarial} & 0.747 \hfill 0.618 & - \qquad - & 0.735 \hfill 0.592 & \textbf{0.891} \hfill \underline{0.644} & 0.811 \hfill 0.623 & 0.850 \hfill \textbf{0.657} & \underline{0.873} \hfill 0.627 \\
          
        \hline
        \multirow{4}{*}{Faster RCNN}
          & Clean                               & 0.963 \hfill \textbf{0.722} & 0.886 \hfill 0.650 & 0.942 \hfill 0.702 & 0.953 \hfill 0.692 & 0.960 \hfill 0.713 & \underline{0.964} \hfill \underline{0.722} & \textbf{0.965} \hfill 0.705 \\
          \cline{2-9}
          & AdvPatch~\cite{thys2019fooling}     & 0.510 \hfill 0.546 & 0.184 \hfill 0.152 & 0.597 \hfill 0.557 & 0.842 \hfill 0.624 & 0.852 \hfill 0.663 & \underline{0.914} \hfill \textbf{0.692} & \textbf{0.932} \hfill \underline{0.682} \\
          & NatPatch~\cite{hu2021naturalistic}  & 0.713 \hfill 0.629 & 0.274 \hfill 0.212 & 0.732 \hfill 0.620 & 0.777 \hfill 0.595 & 0.750 \hfill 0.627 & \underline{0.901} \hfill \textbf{0.683} & \textbf{0.906} \hfill \underline{0.654} \\
          & AdvTexture~\cite{hu2022adversarial} & 0.598 \hfill 0.657 & 0.547 \hfill 0.437 & 0.574 \hfill 0.620 & \underline{0.836} \hfill 0.660 & 0.686 \hfill 0.662 & \textbf{0.888} \hfill \textbf{0.699} & 0.818 \hfill \underline{0.662} \\
          
        \hline
        \multirow{4}{*}{DETR}
          & Clean                               & 0.945 \hfill \underline{0.737} & 0.916 \hfill 0.668 & 0.929 \hfill 0.728 & 0.939 \hfill 0.727 & \underline{0.949} \hfill \textbf{0.740} & 0.944 \hfill 0.735 & \textbf{0.954} \hfill 0.729 \\
          \cline{2-9}
          & AdvPatch~\cite{thys2019fooling}     & 0.656 \hfill 0.659 & 0.410 \hfill 0.325 & 0.683 \hfill 0.633 & 0.857 \hfill 0.691 & 0.861 \hfill 0.695 & \underline{0.889} \hfill \textbf{0.718} & \textbf{0.932} \hfill \underline{0.711} \\
          & NatPatch~\cite{hu2021naturalistic}  & 0.877 \hfill \textbf{0.715} & 0.782 \hfill 0.575 & 0.865 \hfill 0.692 & \underline{0.904} \hfill 0.702 & 0.899 \hfill 0.708 & 0.886 \hfill \underline{0.712} & \textbf{0.925} \hfill 0.702 \\
          & AdvTexture~\cite{hu2022adversarial} & 0.871 \hfill \textbf{0.722} & 0.784 \hfill 0.582 & 0.857 \hfill 0.698 & 0.887 \hfill 0.701 & \underline{0.897} \hfill \underline{0.718} & 0.891 \hfill 0.715 & \textbf{0.909} \hfill 0.715 \\
        \hline
        \hline
        \multirow{2}{*}{Overall} & Clean & 0.939 \hfill \textbf{0.718} & 0.901 \hfill 0.659 & 0.924 \hfill 0.701 & 0.942 \hfill 0.696 & \underline{0.943} \hfill 0.713 & 0.939 \hfill \underline{0.715} & \textbf{0.951} \hfill 0.705 \\
          & Under Attacks & 0.686 \hfill 0.627 & 0.497 \hfill 0.381 & 0.701 \hfill 0.610 & 0.834 \hfill 0.637 & 0.812 \hfill 0.658 & \underline{0.872} \hfill \textbf{0.681} & \textbf{0.893} \hfill \underline{0.667} \\
        \hline
    \end{tabular}
\end{table*}

\subsection{Real-world Validation}
\textbf{Dataset.} Since the APRICOT dataset is already a real-world benchmark for testing creating attacks, here we focus on real-world validation for defending hiding attacks. To this end, we physically printed the AdvPatch trained on the INRIA-Person benchmark and deployed it in the real world, targeting two victim detectors: YOLOv3 and Faster R-CNN.
We set the printed patch size to 12$\times$12 inches, and attached it to a person for testing. Multiple videos were recorded in different environmental conditions: indoor, outdoor lighting (Outdoor-L), and outdoor shadow (Outdoor-S). Importantly, personally identifiable information (PII) was removed from these videos. We selected 600 frames in total from these videos to test the performance of victim detectors, with three different defense methods: NutNet~\cite{lin2024don}, PAD~\cite{jing2024pad}, and \ProjectName{}. 

\textbf{Evaluation metrics.} We use average ASR as our primary evaluation metric here, because it is easier to interpret and directly reflects whether an attack hides the target object. Unlike mAP or AR, ASR allows straightforward human verification of success or failure in real world. Given that every individual in our dataset is clearly visible, we expect the detector to assign high confidence scores to such clearly detectable objects. Therefore, we set the confidence score threshold as 0.9 when calculating ASR. An attack is considered successful if no predicted bounding box has both a confidence score above 0.9 and an IoU greater than 0.5 with the ground-truth bounding box of the targeted person. All evaluation results are cross-validated by two co-authors of this paper to ensure accuracy, with a Cohen's Kappa agreement score of 1 (total agreement).

\section{Evaluation Results}
We evaluate the defense performance of \ProjectName{} from four key aspects: \textit{effectiveness} across diverse attacks and detectors, \textit{efficiency} in processing images, \textit{robustness} against adaptive attacks, and \textit{ablation studies} of its main parameters. For clarity, we use the notation ``Detector-Attack'' to specify an attack method targeting a particular detector. For example, YOLOv3-AdvPatch indicates implementing AdvPatch attack against YOLOv3 object detector.

\subsection{Effectiveness Analysis}
\subsubsection{Defending Hiding Attacks}
Table~\ref{tab:inria} shows the quantitative performance of \ProjectName{} compared to baselines under various hiding attacks. For each row, the best method is marked in \textbf{bold} text, while the second best defense method is marked with an \underline{underline}. The Clean rows indicate the performance on clean images without any attack, while the Vanilla column represents the results of detectors without any defense applied. We also show the overall performance of each method in the bottom block. From the Vanilla column, we can observe that all object detectors are vulnerable to three adversarial attacks, where transformer-based DETR is more robust than CNN-based YOLOv3 and Faster RCNN.
Regarding defense methods, ObjectSeeker (OS) is a certifiable defense method with a threat model that assumes only one single adversarial patch is present within an image, its performance degrades significantly in our experiments as most images contain multiple adversarial patches. 
Among the empirical defense methods, SAC can only improve the performance of victim detectors under AdvPatch attack, but lacks generalization to the other two attacks NatPatch and AdvTexture. 
Jedi and NutNet show similar overall performance, with Jedi performing better on mAP@0.5, while NutNet performs better in terms of AR. PAD obtains the best performance among all the other baselines when under attacks, with overall mAP@0.5 of 0.872 and AR of 0.681. Our method \ProjectName{} achieves the best overall mAP@0.5 of 0.893 when facing attacks, 2.1\% better than PAD, and the second best AR of 0.667. 
For clean images, the proposed \ProjectName{} achieves even better mAP@0.5 score than the Vanilla results, with a slight drop on AR. \revise{The stronger mAP@0.5 on clean data likely reflects improved data distribution alignment: the regenerated images more closely match the detector’s training data distribution. Since both the LDM and the object detectors were trained on MS-COCO, regeneration can ``pull'' the images toward COCO distribution and thus improve detection performance. Overall, t}hese results demonstrate that \ProjectName{} achieves superior performance against hiding attacks, and preserves utility on clean images.

\begin{figure*}
    \centering
    \includegraphics[width=0.95\textwidth]{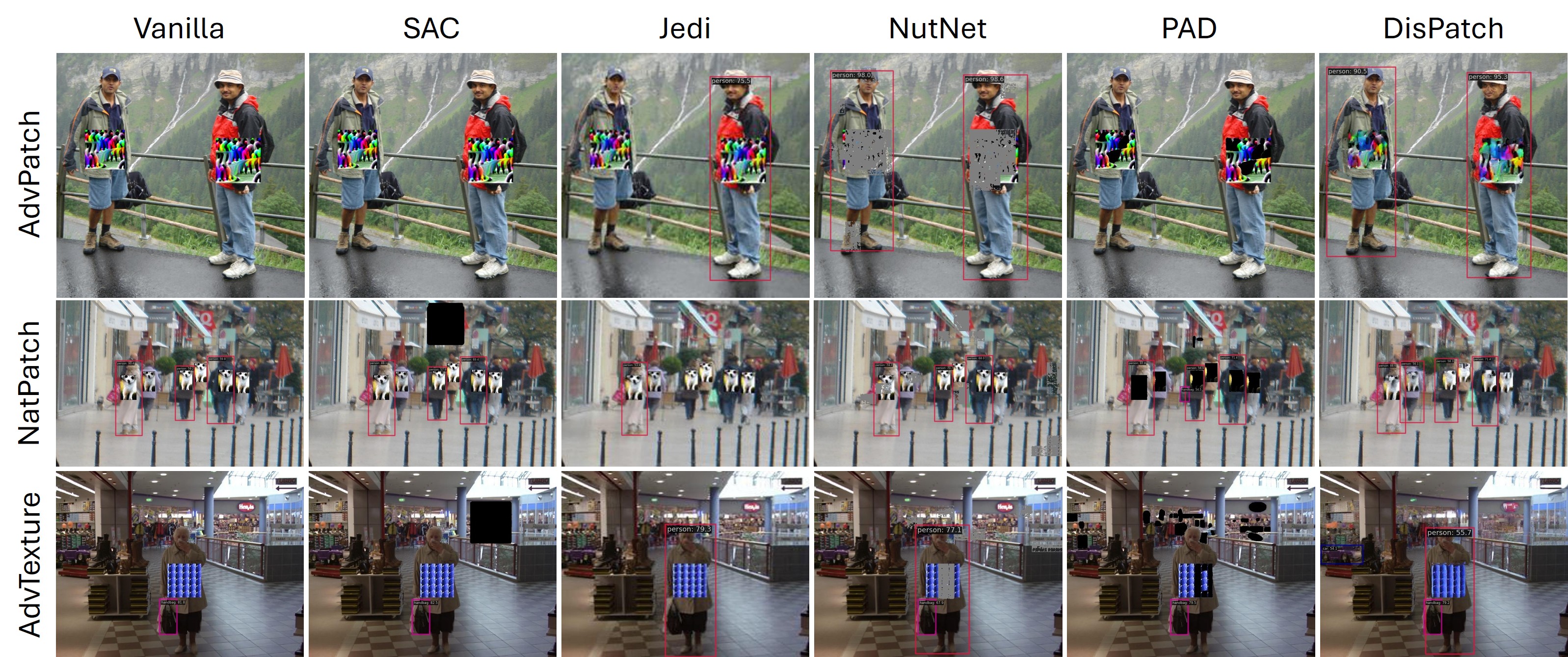}
    \caption{Qualitative results of our \ProjectName{} compared with other baseline methods when defending YOLOv3 against AdvPatch, NatPatch, and AdvTexture attacks on INRIA-Person dataset. Red bounding boxes indicate the detected persons.}
    \label{fig:yolov3_hiding}
\end{figure*}

\textbf{Qualitative results.} Figure~\ref{fig:yolov3_hiding} shows sample results of our \ProjectName{} method compared to the other baselines in defending YOLOv3 against AdvPatch, NatPatch, and AdvTexture. Additional results are provided in public code repository. Without any defense (Vanilla), all of the three types of attacks successfully hide at least one person in the demo images. 
SAC incorrectly identifies some benign regions as adversarial patches in the demo images, while Jedi is able to recover several missed detections but simultaneously causes other correct detections to fail.
NutNet successfully detects adversarial patches in both AdvPatch and AdvTexture examples, but fails to identify the patches in the NatPatch example. PAD removes a small portion of the adversarial patches in the AdvPatch example and AdvTexture example, leading to two unsuccessful defenses. But it can detect all patches in NatPatch example. In contrast,  \ProjectName{} successfully replaces the adversarial patches with benign counterparts, enabling correct detection in AdvPatch and AdvTexture examples. Additionally, in the NatPatch example, \ProjectName{} recovers one previously undetected person. Although patch-like patterns may still be visually noticeable in the outputs of \ProjectName{}, these regions no longer retain adversarial functionality. 



\begin{table*}[h]
    \centering
    \caption{The performance of \ProjectName{} compared with other defense methods against creating attacks on APRICOT benchmark. In this table, the lower values indicate better defense performance. Here, mAP refers to mAP@0.5.}
    \label{tab:apricot}
    \begin{tabular}{|c|c|c|c|c|c|c|c|}
        \hline
        \multirow{2}{*}{Detector} & \multirow{2}{*}{Metrics} & No Defense & \multicolumn{5}{c|}{Defense} \\
        \cline{3-8}
         &  & Vanilla & SAC~\cite{liu2022segment} & Jedi~\cite{tarchoun2023jedi} & NutNet~\cite{lin2024don} & PAD~\cite{jing2024pad} & \ProjectName{} \\ 
         &  & mAP \hfill ASR & mAP \hfill ASR & mAP \hfill ASR & mAP \hfill ASR & mAP \hfill ASR &  mAP \hfill ASR \\ 
        \hline
        \multirow{2}{*}{Faster RCNN} 
          & Targeted & 0.123 \hfill 0.171 & 0.028 \hfill 0.037 & 0.026 \hfill \underline{0.030} & 0.051 \hfill 0.067 & \underline{0.025} \hfill 0.037 & \textbf{0.018} \hfill \textbf{0.017} \\
          & Untargeted & 0.355 \hfill 0.589 & 0.164 \hfill 0.431 & 0.197 \hfill 0.415 & \textbf{0.087} \hfill \underline{0.338} & 0.158 \hfill 0.398 & \underline{0.114} \hfill \textbf{0.327} \\
        \hline
        \multirow{2}{*}{SSD}
          & Targeted & 0.012 \hfill 0.017 & 0.013 \hfill 0.017 & 0.036 \hfill 0.049 & 0.007 \hfill \underline{0.007} & \underline{0.005} \hfill \underline{0.007} & \textbf{0.004} \hfill \textbf{0.003} \\
          & Untargeted & 0.111 \hfill 0.315 & 0.181 \hfill 0.437 & 0.064 \hfill 0.231 & \textbf{0.014} \hfill \underline{0.178} & 0.097 \hfill 0.301 & \underline{0.015} \hfill \textbf{0.171} \\
        \hline
        \multirow{2}{*}{RetinaNet}
          & Targeted & 0.043 \hfill 0.059 & 0.029 \hfill 0.035 & 0.013 \hfill 0.021 & 0.017 \hfill 0.024 & \textbf{0.000} \hfill \textbf{0.007} & \underline{0.009} \hfill \underline{0.019} \\
          & Untargeted & 0.156 \hfill 0.368 & 0.125 \hfill 0.365 & 0.170 \hfill 0.365 & \textbf{0.050} \hfill \textbf{0.233} & 0.097 \hfill 0.316 & \underline{0.080} \hfill \underline{0.247} \\
        \hline
        \hline
        \multirow{2}{*}{Overall}
          & Targeted & 0.059 \hfill 0.082 & \underline{0.023} \hfill 0.030 & 0.025 \hfill 0.033 & 0.025 \hfill 0.033 & \textbf{0.010} \hfill \underline{0.017} & \textbf{0.010} \hfill \textbf{0.013} \\
          & Untargeted & 0.207 \hfill 0.424 & 0.157 \hfill 0.411 & 0.144 \hfill 0.337 & \textbf{0.050} \hfill \underline{0.250} & 0.117 \hfill 0.338 & \underline{0.070} \hfill \textbf{0.248} \\
        \hline
    \end{tabular}
\end{table*}

\subsubsection{Defending Creating Attacks}
Table~\ref{tab:apricot} shows the performance of different defense methods on the APRICOT benchmark. The overall ASR of targeted attacks on Vanilla detectors is 0.082, and of untargeted attacks is 0.424, showing that untargeted attacks work much better than targeted attacks. These results are consistent with original results reported in APRICOT paper~\cite{braunegg2020apricot}. Among all the other baselines, PAD performs the best when defending targeted attacks, with an overall mAP@0.5 of 0.010 and ASR of 0.017. On the other hand, NutNet outperforms the other baselines when defending untargeted attacks, with the lowest overall mAP@0.5 of 0.050. Our method \ProjectName{} achieves the lowest ASRs on Faster RCNN and SSD when defending both targeted and untargeted attacks, and the second lowest on RetinaNet. Overall, \ProjectName{} shows the best performance when facing targeted attacks, and competitive performance as NutNet when facing untargeted attacks. By combining results from Table~\ref{tab:inria} and Table~\ref{tab:apricot}, we can observe that \ProjectName{} demonstrates consistent superior performance across different types of attacks.

\begin{figure*}
    \centering
    \includegraphics[width=0.95\textwidth]{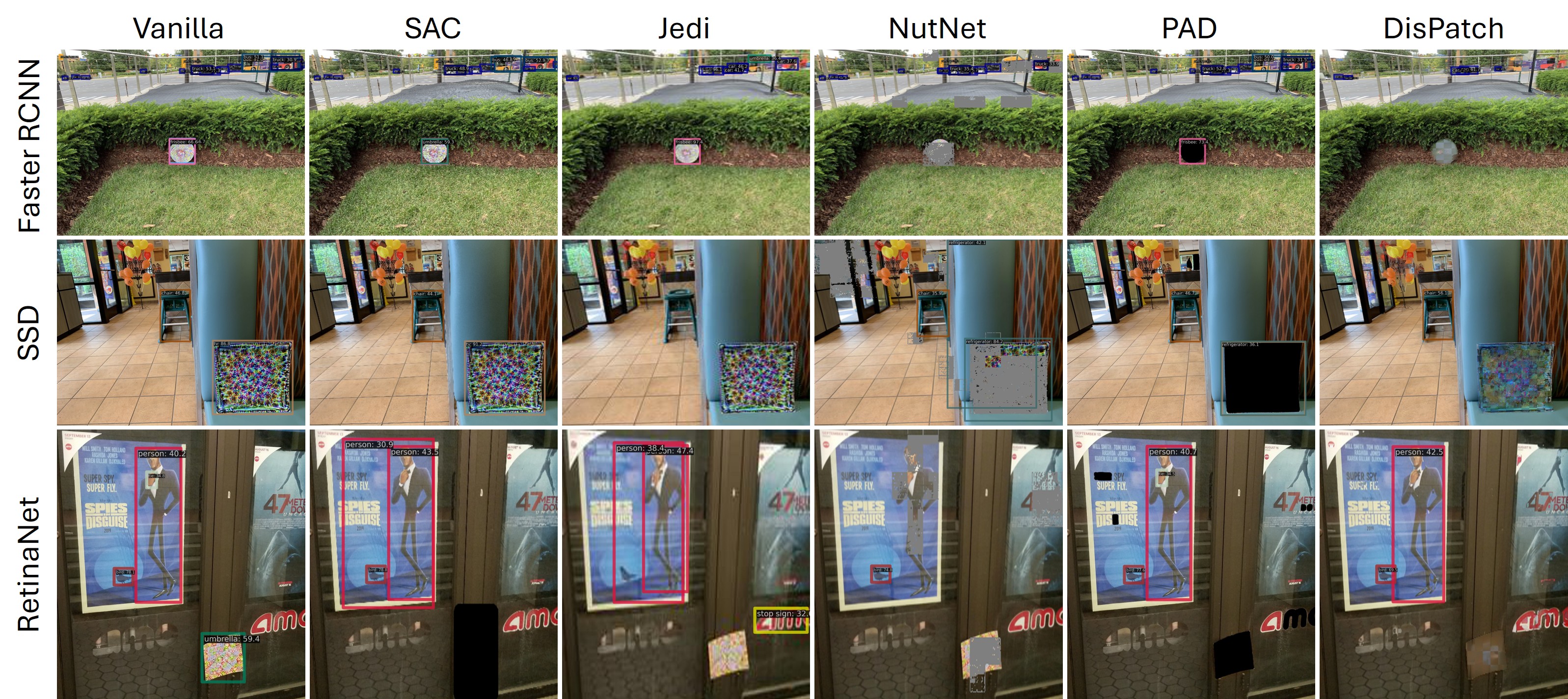}
    \caption{Qualitative results of our \ProjectName{} compared with the baseline methods when defending Faster RCNN, SSD, and RetinaNet on APRICOT dataset.}
    \label{fig:creating}
\end{figure*}

\textbf{Qualitative results.} We show the qualitative results of \ProjectName{} compared to the other baselines for defending creating attacks in Figure~\ref{fig:creating}. The top-to-bottom rows show attacks on Faster RCNN, SSD, and RetinaNet, respectively.  In all three examples, the adversarial patches successfully launched creating attacks. SAC can only locate and remove the adversarial patch in the RetinaNet example, while Jedi does not find any adversarial patches in these examples. NutNet successfully detects all patches in the examples but also masks out some benign regions. PAD is also capable of identifying all adversarial patches, however, it replaces them with black pixels. These blacked-out regions are still recognized as objects by detectors such as Faster RCNN and SSD, leading to false positive detections. The proposed \ProjectName{} disarms adversarial patches in all examples with regenerated benign content, showing its effectiveness against these noise-like creating attacks. Upon analyzing these examples, we observe that existing ``detect and remove'' baselines have the risks of either removing benign regions or failing to eliminate non-existent objects, while \ProjectName{} can reduce these risks through our ``regenerate and rectify'' approach.



\subsubsection{Real-world Validation} Experimental results of real-world validation are shown in Table~\ref{tab:realworld}, comparing the performance of \ProjectName{} with two strong baselines NutNet and PAD. The ASRs of AdvPatch are high on the Vanilla detectors, with over 80\% on YOLOv3 and over 35\% on Faster RCNN. Across all environmental conditions, \ProjectName{} substantially reduces ASR to low values, consistently outperforming NutNet, and only underperforming PAD in the outdoor shadow condition for YOLOv3. However, PAD exhibits unstable performance across different conditions, with much higher ASRs under both indoor and outdoor lighting conditions for YOLOv3. Overall, these results demonstrate that \ProjectName{} can effectively mitigate the impact of adversarial patches in real-world scenarios, and restore the confidence scores of victim objects to high levels (above 0.9). 

\begin{table}[]
    \centering
    \caption{ASRs of the AdvPatch attack in real-world tests under different conditions (L: lighting; S: shadow).}
    \resizebox{88mm}{!}{
    \begin{tabular}{|c|c|c|c|c|c|c|}
    \hline
    Detectors & \multicolumn{3}{c|}{YOLOv3} & \multicolumn{3}{c|}{Faster RCNN} \\
    \hline
        Condition & Indoor & Outdoor-L & Outdoor-S & Indoor & Outdoor-L & Outdoor-S \\
    \hline
        Vanilla & 0.99 & 0.80 & 1.00 & 0.42 & 0.43 & 0.36 \\
    \hline
        NutNet~\cite{lin2024don} & 0.24 & 0.23 & 0.27 & 0.09 & 0.13 & 0.03  \\
        PAD~\cite{jing2024pad} & 0.39 & 0.40 & \textbf{0.21} & 0.04 & 0.11 & 0.00  \\
        \ProjectName{} & \textbf{0.20} & \textbf{0.17} & 0.26 & \textbf{0.02} & \textbf{0.09} & \textbf{0.00} \\
    \hline
    \end{tabular}
    }
    \label{tab:realworld}
\end{table}

\begin{figure}[h]
    \centering
    \includegraphics[width=0.95\linewidth]{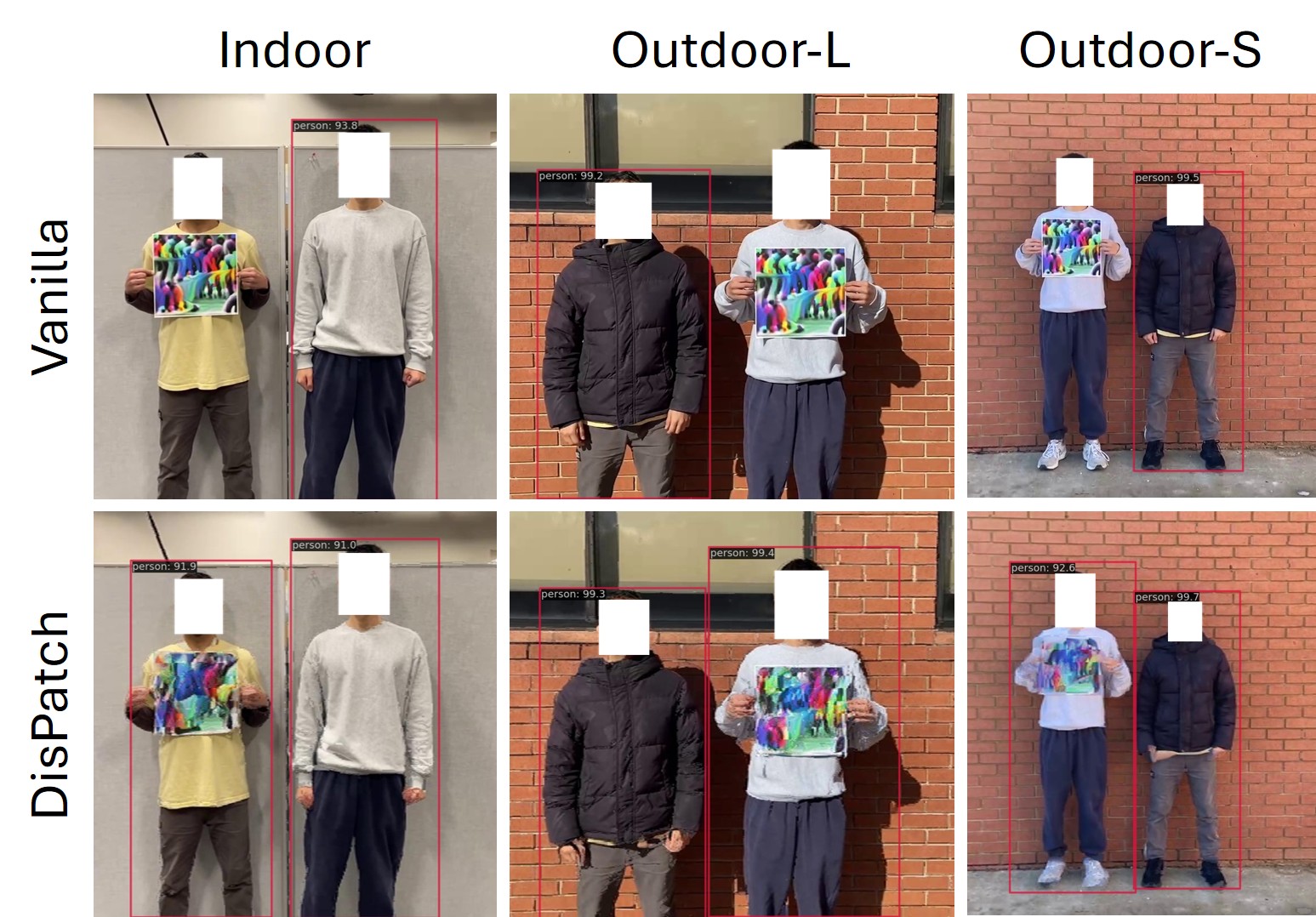}
    \caption{Qualitative results of real world validation. \ProjectName{} successfully restores the person hidden by the patch, regardless of the environmental conditions.}
    \label{fig:realworld}
\end{figure}

\textbf{Qualitative results.} Figure~\ref{fig:realworld} presents qualitative results of \ProjectName{} on our real-world dataset across various environmental conditions, including indoor, outdoor lighting (Outdoor-L), and outdoor shadow (Outdoor-S). As illustrated, \ProjectName{} effectively restores individuals hidden by adversarial patches in diverse scenarios, demonstrating its practical applicability in real-world environments.

\begin{figure}
    \centering
    \includegraphics[width=\linewidth]{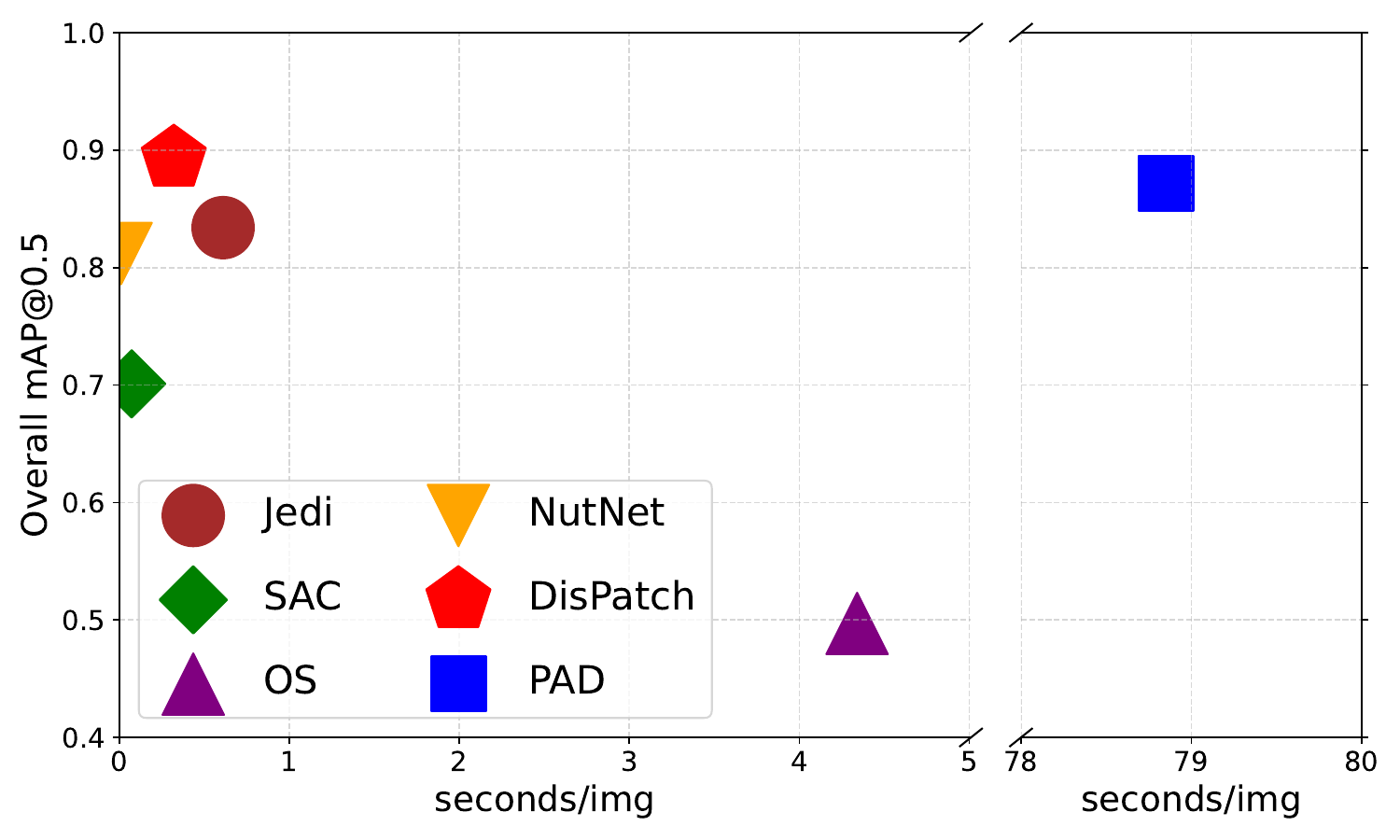}
    \caption{The trade-off figure between efficiency (seconds/img) and effectiveness (Overall mAP@0.5).}
    \label{fig:tradeoff}
\end{figure}

\revise{\textit{\textbf{Takeaways:} Through the effectiveness analysis, \ProjectName{} achieves consistent performance on both hiding and creating attacks. In contrast, PAD performs well on hiding attacks but degrades on creating attacks, while NutNet performs well on creating attacks but is less effective on hiding attacks. Overall, these results show that \ProjectName{} generalizes better across different attack types.}}

\subsection{Efficiency Analysis}\label{sec:exp_time} 
The efficiency of defense algorithms is a critical factor for their deployment in real-world systems. In Figure~\ref{fig:tradeoff}, we present a trade-off plot between the efficiency and effectiveness of various defense methods against hiding attacks on INRIA-Person. The X-axis represents the average processing time for one input image (excluding the detector processing time), while the Y-axis shows the average mAP@0.5 across all hiding attacks and detectors. An ideal defense algorithm would have high mAP@0.5 performance and short processing time, placing it in the upper-left region of the figure. As observed, \ProjectName{} outperforms all other baselines in terms of effectiveness, while maintaining high efficiency. Although PAD achieves the second best result on mAP@0.5, its average processing time for each image exceeds one minute, falls behind all the other methods. \ProjectName{} completes the regeneration stage in 0.28 seconds and the rectification stage in 0.04 seconds per image. Whereas \ProjectName{} is not the fastest method, it strikes a balance between effectiveness and efficiency. This makes it particularly well-suited for offline processing applications where real-time constraints are relaxed, such as post-event surveillance video analysis, or batch processing in security systems. In such contexts, the slightly longer processing time is a worthwhile trade-off for the significantly enhanced defense performance.



\subsection{Robustness Analysis}\label{sec:exp_adaptive}

An important aspect of an adversarial defense algorithm is its robustness against future adaptive attacks. Adaptive attacks refer to scenarios in which attackers adjust their strategy based on knowledge of the deployed defense mechanisms. In our setting, we assume that the attacker is aware of both the \ProjectName{} defense and the victim object detector, and thus constructs a new adversarial attack specifically designed to evade the defense.

To simulate an adaptive attack against the \ProjectName{} defense, the attacker can add an additional term to the loss function for optimizing the patch, alongside the original attack loss function:
\begin{equation}
    \mathcal{L} = \mathcal{L}_{attack}  + \alpha\mathcal{L}_{DisPatch}
\end{equation}
where $\mathcal{L}_{attack}$ is the original attack loss function, $\mathcal{L}_{DisPatch}$ is the loss function for evading \ProjectName{}, and $\alpha$ controls the importance of $\mathcal{L}_{DisPatch}$. Since \ProjectName{} uses the L2 distance between the original input image $I$ and the regenerated image $\tilde{I}$ to identify adversarial regions, we define $\mathcal{L}_{DisPatch}$ as the average pixel L2 distance between a patch $\delta$ and its regenerated counterpart $\tilde{\delta}$:
\begin{equation}
    \mathcal{L}_{DisPatch} = \mu(\parallel \delta - \tilde{\delta}\parallel_2 )
\end{equation}
where $\mu$ represents the average over all pixels within $\delta$. In this way, we can force the adversarial patch to retain its original pattern even after undergoing the diffusion regeneration stage, thereby increasing its chances of evading \ProjectName{}. We experiment with four different values of $\alpha$: 0, 0.5, 1, and 5.0, to launch adaptive YOLOv3-AdvPatch attacks and YOLOv3-NatPatch attacks on INRIA-Person dataset, where $\alpha=0$ corresponds to a non-adaptive attack baseline.

The results of adaptive attacks are shown in Table~\ref{tab:adaptive}. For each adaptive attack, we report the mAP@0.5 performance of the victim object detector, without any defense (Vanilla) or with \ProjectName{} defense. As we can see from the table, non-adaptive AdvPatch attack and NatPatch attacks reduce the mAP@0.5 of vanilla YOLOv3 to 0.498 and 0.707, respectively. Adding $\mathcal{L}_{DisPatch}$ to the loss function significantly weakens the attack capability of both AdvPatch and NatPatch attacks. When $\alpha=0.5$, vanilla YOLOv3 obtains much better performance, with 0.761 under adaptive AdvPatch attack and 0.765 under adaptive NatPatch attack. As $\alpha$ increases, the attack capability continues to decline, since the optimization process shifts focus toward evading the defense rather than attacking the victim detector. However, these adaptive attacks still fail to evade the defense of \ProjectName{}. With \ProjectName{} applied, the detection performance of the victim detector is restored, reaching mAP@0.5 scores higher than 90\% under adaptive AdvPatch attack, and over 84\% under adaptive NatPatch attack. 

\begin{table}[]
    \centering
    \caption{mAP@0.5 results of YOLOv3 on INRIA-Person under different adaptive attacks, where $\alpha=0$ means non-adaptive attack.}
    \label{tab:adaptive}
    \resizebox{\linewidth}{!}{
    \begin{tabular}{|c|c|c|c|c|c|}
    \hline
         Attacks & Defense & $\alpha=0$ & $\alpha=0.5$ & $\alpha=1.0$ & $\alpha=5.0$ \\
    \hline
        \multirow{2}{*}{AdvPatch} & Vanilla & 0.498 & 0.761 & 0.802 & 0.857  \\
    \cline{2-6}
         & \ProjectName{} & 0.903 & 0.908 & 0.908 & 0.904  \\
    \hline
        \multirow{2}{*}{NatPatch} & Vanilla & 0.707 & 0.765 & 0.808 & 0.826 \\
    \cline{2-6}
         & \ProjectName{} & 0.840 & 0.845 & 0.864 & 0.868 \\
    \hline
    \end{tabular}
    }
\end{table}
\begin{figure}[]
    \centering
    \includegraphics[width=\linewidth]{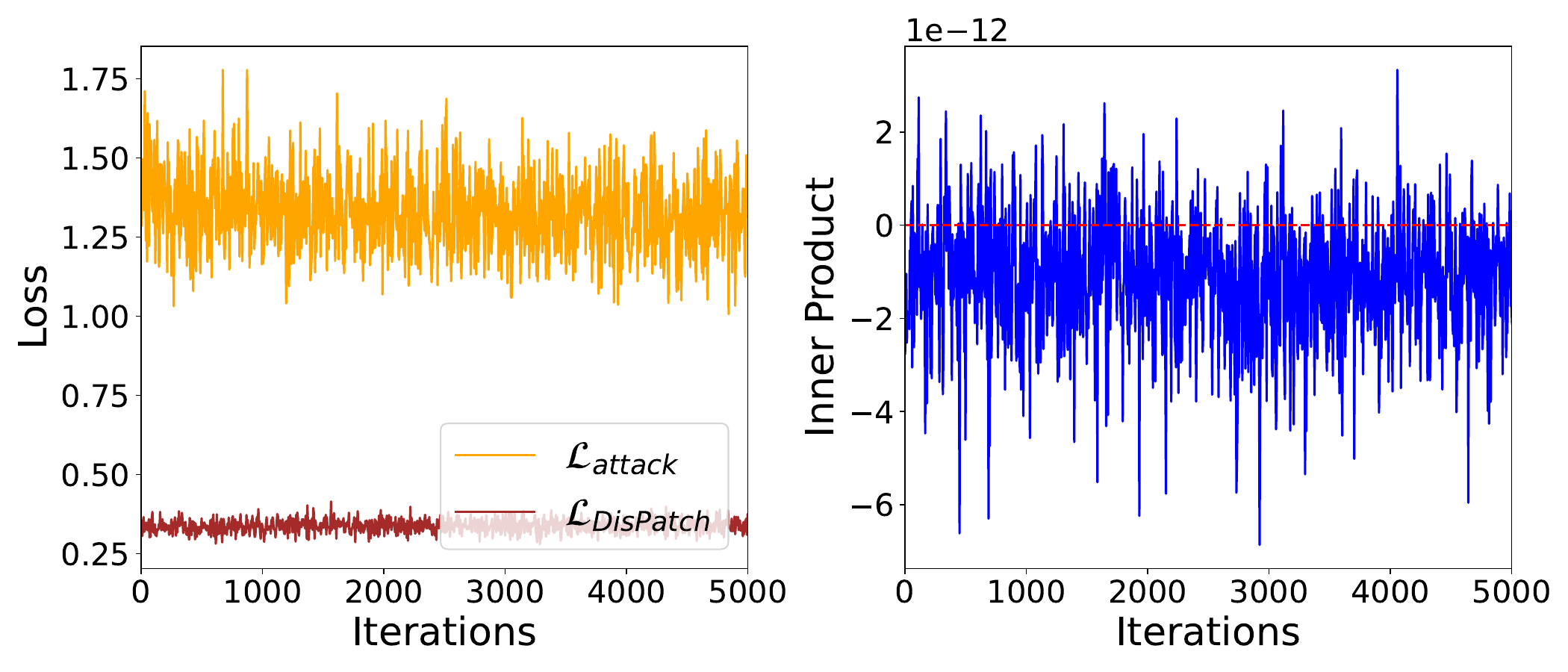}
    \caption{Left: $\mathcal{L}_{DisPatch}$ and $\mathcal{L}_{attack}$ losses over iterations. Right: the inner product of their gradients over iterations. }
    \label{fig:losses}
\end{figure}

\begin{figure}[t]
    \centering
    \includegraphics[width=\linewidth]{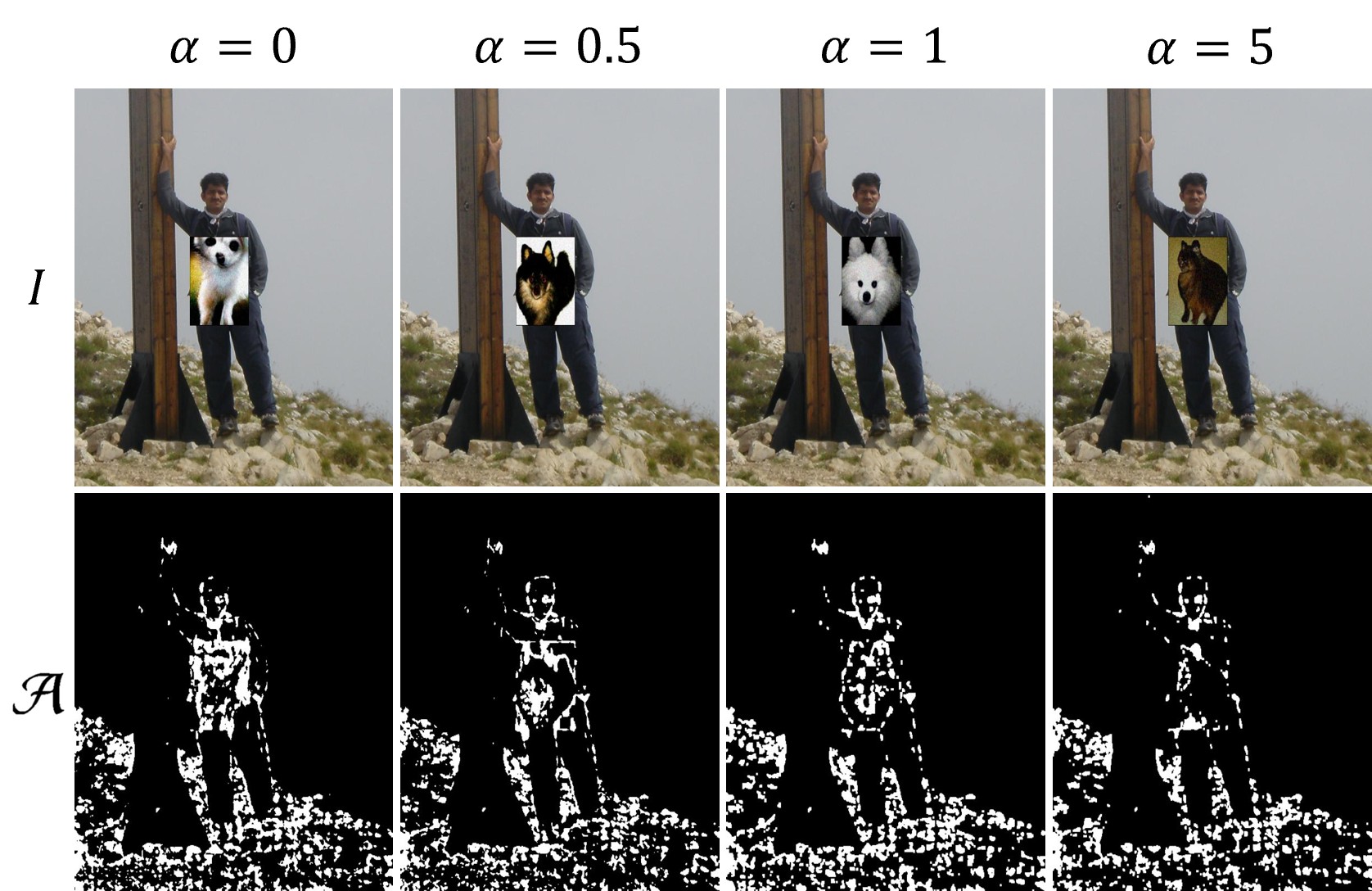}
    \caption{Qualitative results of adaptive YOLOv3-NatPatch attacks with different $\alpha$. Top row: attacked images, bottom row: adversarial masks detected by \ProjectName{}.}
    \label{fig:adaptive}
\end{figure}

To further analyze the reason behind this phenomenon, we visualize the behavior of the two loss functions, $\mathcal{L}_{attack}$ and $\mathcal{L}_{DisPatch}$, during the training of an adaptive AdvPatch attack. Figure~\ref{fig:losses} shows the loss values over iterations, along with the inner product of their gradients. As illustrated, it is hard to optimize these two losses $\mathcal{L}_{attack}$ and $\mathcal{L}_{DisPatch}$ at the same time, as they do not converge even after 5000 iterations. Moreover, the right subplot reveals that the inner product between their gradients remains negative throughout almost all iterations, indicating that \textit{their optimization directions are largely contradictory}. This finding aligns with the observation from NutNet~\cite{lin2024don}, and explains why these two losses do not converge over iterations. We show some examples of adaptive YOLOv3-NatPatch attacked images and their corresponding adversarial masks $\mathcal{A}$ detected by \ProjectName{} in Figure~\ref{fig:adaptive}. As the value of $\alpha$ increases, the adaptive attacks become more effective at evading \ProjectName{}, where fewer adversarial regions are detected. However, this comes at the cost of reduced attack strength, as shown in Table~\ref{tab:adaptive}. Despite these evasions in some cases, the overall performance of \ProjectName{} remains stable, demonstrating its robustness against adaptive attacks.

\subsection{Ablation Studies} \label{sec:eval_ablation}
\subsubsection{Number of Grids $N$} 
\begin{figure}
    \centering
    \includegraphics[width=\linewidth]{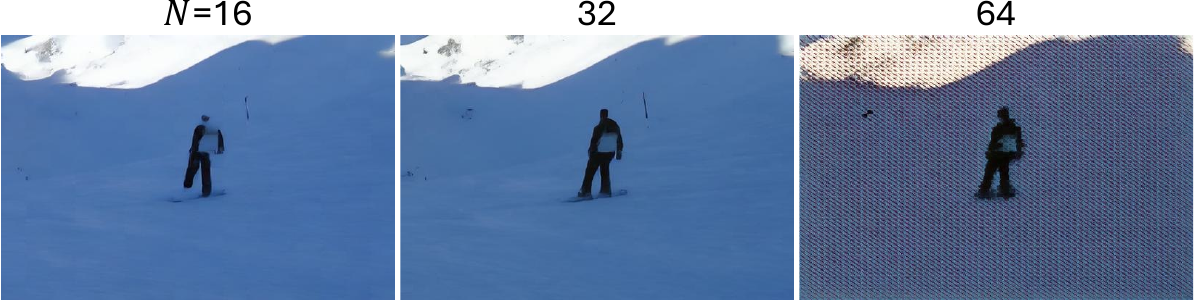}
    \caption{\revise{The regenerated images $\tilde{I}$ with different values of $N$.}}
    \label{fig:ablation_N}
\end{figure}

\revise{Our default choice of $N=32$ is guided by an empirical trade-off that each grid cell should be small enough to avoid covering an entire object, especially small objects. In MS-COCO dataset~\cite{lin2014microsoft}, a small object is defined as having area below $32\times32$ pixels in a typical $640\times480$ image, \ie, roughly $\frac{1}{20}\times\frac{1}{15}$ of the image in width and height. With $N=32$, each grid cell spans $\frac{1}{32}\times\frac{1}{32}$ of the image, which is smaller than this small-object scale and thus reduces the chance that a single cell fully contains a small object.} 
\revise{To observe the impact of $N$ to the defense performance, w}e change the values of $N$ and report the mAP@0.5 performance of \ProjectName{} for different hiding attacks on INRIA-Person in Table~\ref{tab:num_grids_hiding}\revise{, and show one qualitative example in Figure~\ref{fig:ablation_N}}. As we can see from the table, compared to the default $N=32$, the smaller value of $N=16$ generally leads to performance degradation of at most 5\%. This degradation occurs because a smaller $N$ results in larger grid sizes in the inpainting masks during the regeneration stage. Some masked grids may cover small objects, causing the inpainting method to inadvertently remove these objects. \revise{For example, when $N=16$ in Figure~\ref{fig:ablation_N}, the person’s head and feet, as well as the snowboard, are noticeably degraded.} A larger $N=64$ leads to much worse performance of \ProjectName{}, because it makes the grid size too small to provide sufficient context information during the regeneration stage. \revise{In addition, the dense masking pattern also harms generation quality, producing a visible checkerboard artifact as shown in Figure~\ref{fig:ablation_N}. Overall, we find that $N=32$ provides the best balance across different scenarios.} 

\begin{table}[]
    \centering
    \caption{The impact of $N$ on \ProjectName{} performance (mAP@0.5) on INRIA-Person dataset.}
    \label{tab:num_grids_hiding}
    \begin{tabular}{|c|c|c|c|c|}
    \hline
      Detector & $N$ & 16 & 32 & 64 \\
    \hline
      \multirow{4}{*}{YOLOv3} 
       & Clean & 0.929 & 0.934 & 0.875 \\
       \cline{2-5}
       & AdvPatch & 0.888 & 0.903 & 0.747 \\
       & NatPatch & 0.822 & 0.840 & 0.646 \\
       & AdvTexture & 0.823 & 0.873 & 0.734 \\
    \hline
      \multirow{4}{*}{Faster RCNN} 
       & Clean & 0.950 & 0.965 & 0.885 \\
       \cline{2-5}
       & AdvPatch & 0.914 & 0.932 & 0.751 \\
       & NatPatch & 0.890 & 0.906 & 0.702 \\
       & AdvTexture & 0.767 & 0.814 & 0.580 \\
    \hline
      \multirow{4}{*}{DETR} 
       & Clean & 0.945 & 0.954 & 0.850 \\
       \cline{2-5}
       & AdvPatch & 0.904 & 0.932 & 0.756 \\
       & NatPatch & 0.911 & 0.925 & 0.754 \\
       & AdvTexture & 0.863 & 0.909 & 0.764 \\
    \hline
    \end{tabular}
\end{table}

\subsubsection{Sampling Steps $s$} 
\begin{figure}[!b]
    \centering
    \includegraphics[width=\linewidth]{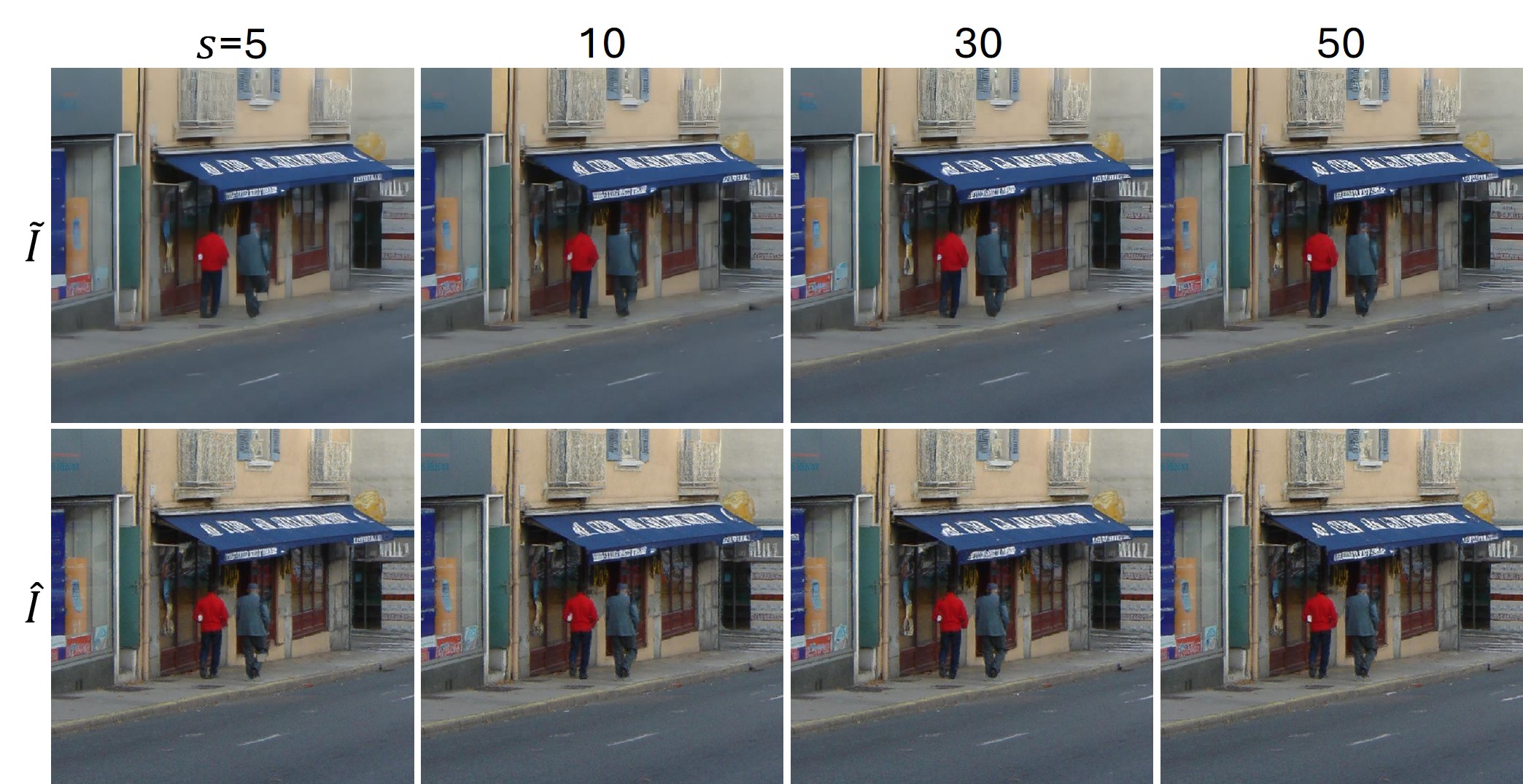}
    \caption{The regenerated image $\tilde{I}$ and \ProjectName{} output image $\hat{I}$ with different sampling steps $s$. 
    }
    \label{fig:sampling}
\end{figure}

The number of sampling steps $s$ in diffusion models plays a crucial role in balancing output quality and computational efficiency. Generally, increasing $s$ leads to higher quality image generation, but incurs greater inference time, whereas reducing $s$ accelerates processing, making the model more suitable for real-time applications. In \ProjectName{}, we use a default setting of $s=5$ to prioritize efficiency. To further understand this trade-off, we evaluate the performance of \ProjectName{} under varying sampling steps using both clean images and YOLOv3-AdvPatch attacked images, as shown in Table~\ref{tab:time}. Interestingly, we observe that increasing the number of sampling steps does not lead to better performance. We hypothesize that with fewer sampling steps, the regenerated image exhibits greater divergence from adversarial regions, which in turn enhances the contrast between benign and perturbed areas. Some visual examples with different sampling steps are shown in Figure~\ref{fig:sampling}. As illustrated in the figure, increasing the number of sampling steps enhances the preservation of details in the generated image $\tilde{I}$. For instance, the legs of the person on the right become more discernible. However, this improvement comes at the cost of increased inference time. In conclusion, the small value of $s=5$ is the optimal choice for \ProjectName{}.

\begin{table}[]
    \centering
    \caption{The time complexity and mAP@0.5 performance of \ProjectName{} with different diffusion sampling steps $s$, defending YOLOv3 on INRIA-Person dataset.}
    \label{tab:time}
    \begin{tabular}{|c|c|c|c|c|c|}
    \hline
        \multicolumn{2}{|c|}{Sampling steps $s$} & 5 & 10 & 30 & 50 \\
    \hline
        \multicolumn{2}{|c|}{Time (seconds/img)} & 0.32 & 0.67 & 1.95 & 3.16 \\
    \hline
        \multirow{2}{*}{YOLOv3} & Clean & 0.934 & 0.929 & 0.934 & 0.934 \\
         & AdvPatch & 0.903 & 0.896 & 0.899 & 0.901 \\
    \hline
    \end{tabular}
\end{table}

\subsubsection{Adversarial Patch Size}
\begin{figure}
    \centering
    \includegraphics[width=\linewidth]{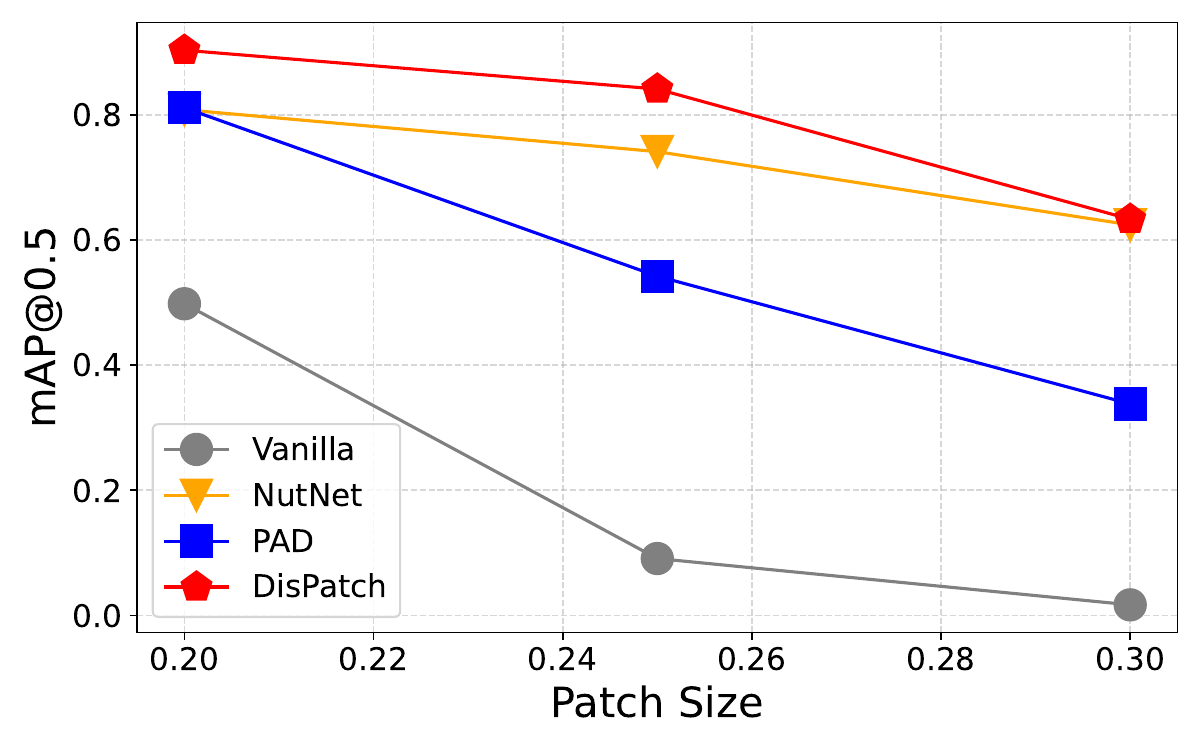}
    \caption{The impact of patch size on the performance of different defense methods against YOLOv3-AdvPatch attack.}
    \label{fig:patch_size}
\end{figure}

The default patch size scale ratio, defined as the ratio of the patch height to the diagonal length of the object bounding box, is set to 0.2 for AdvPatch and NatPatch, and 0.25 for AdvTexture. To evaluate the defense performance of the proposed \ProjectName{} on different sizes of adversarial patches, we modify the patch size scale ratio for the YOLOv3-AdvPatch attack and compare it with two baselines, NutNet and PAD. The results are shown in Figure~\ref{fig:patch_size}. As observed, the mAP@0.5 performance of Vanilla YOLOv3 drops rapidly as the adversarial patch size increases, resulting in less than 10\% when the patch size is 0.3. PAD is sensitive to changes in patch size, it performs better than NutNet at the scale of 0.2, but becomes less effective as the patch size increases. Both NutNet and \ProjectName{} show their robustness against increasing patch size, retaining mAP@0.5 score above 60\% even when patch size is 0.3. Specifically, \ProjectName{} continuously outperforms the other two baselines, showing its stable defense capability.

\subsubsection{Impact to Benign Regions} Previous analyses show that \ProjectName{} can effectively disarms adversarial patches, here we assess its impact on benign regions. We compare \ProjectName{} with two strong baselines, NutNet and PAD, on the INRIA-Person dataset by computing the mean squared error (MSE) between benign regions before and after applying the defense across different attacks. All pixel values are normalized to the range $[0,1]$ in this analysis, and lower MSE values indicate less distortion of benign regions. As shown in Table~\ref{tab:benign_impact}, DisPatch preserves benign regions substantially better than PAD, while performing slightly worse than NutNet. This is expected, as the regeneration process in \ProjectName{} inevitably introduces minor changes in benign areas. However, the detection results in Table~\ref{tab:inria} indicate that these changes have negligible impact on overall object detection performance.

\begin{table}[tb]
    \centering
    \caption{MSE between benign regions before and after applying defenses on INRIA-Person dataset.}
    \label{tab:benign_impact}
    \begin{tabular}{|c|c|c|c|}
    \hline
    Method & NutNet & PAD & \ProjectName{} \\
    \hline
    MSE & 0.0011 & 0.0066 & 0.0024 \\
    \hline
    \end{tabular}
\end{table}
\section{Discussion}\label{sec:discussion}
\textbf{Computational latency.} The main limitation of \ProjectName{} lies in its time complexity, with an average processing time of 0.32 seconds per image. While this may pose challenges for time-critical applications, such as autonomous driving, it is well-suited for offline or near real-time scenarios like surveillance video analysis, forensic review, or batch processing in security pipelines. Importantly, this modest computational cost brings substantial gains in robustness and defense effectiveness. Furthermore, \ProjectName{} offers clear opportunities for future optimization, such as adopting lightweight diffusion architectures, model pruning, or quantization, paving the way for broader deployment across various operational settings.

\revise{\textbf{Real-world implications.} We highlight a concrete deployment scenario where \ProjectName{} is well matched. Since 2024, Transport for London has begun deploying computer vision systems in the London Underground to detect crime and unsafe behaviors~\cite{london2024}, such as the suspected carrying of weapons. The reliability of such systems depends critically on the robustness of their underlying object detectors. A malicious actor could exploit this by applying physical adversarial patches on clothes to evade detection (hiding attacks) or trigger false alarms (creating attacks). In this context, \ProjectName{} can be integrated as a robustness enhancement module in settings where additional latency is acceptable, such as post-event or near real-time analysis of recorded video segments. For example, when investigators need to locate a suspect but the detector produces unreliable results, \ProjectName{} can be applied to the relevant footage to mitigate potential patch effects and improve the detector’s ability to recover evidence-critical detections.}



\textbf{Alternative generative models.} 
\ProjectName{} leverages diffusion models for in-distribution reconstruction, but can other generative models be used? To investigate, we replaced LDM with a Masked Autoencoder (MAE)~\cite{he2022masked} in a small-scale experiment on INRIA-Person with YOLOv3 detector. The MAE-based variant showed a minor drop of 1\% on mAP@0.5 and 2\% on AR, likely due to lower reconstruction quality. These results highlight the adaptability of \ProjectName{}, suggesting that it can be integrated with other generative models. \revise{In real-world deployments, developers can select different generative backbones to match application constraints. For example, lightweight models (\eg, MAE~\cite{he2022masked}) can reduce latency, while larger generative models (\eg, Qwen-Image~\cite{wu2025qwenimagetechnicalreport}) can improve visual fidelity.} We anticipate that as more advanced generative models are developed, the defense performance of \ProjectName{} could be further enhanced.

\textbf{Other potential risks.} Our threat model assumes that the adversary has full knowledge of the architecture and weights of both the victim detector and the defense method, but lacks the capability to tamper with or modify them. This assumption can be reasonably upheld by using models trained in-house or obtained from trusted sources. However, incorporating models or training data from untrusted sources introduces the risk of backdoor attacks~\cite{li2022backdoor}. For instance, if the diffusion model in \ProjectName{} was compromised via a backdoor, an attacker could exploit hidden triggers to manipulate the generated image, potentially rendering the defense ineffective. Although such scenarios fall outside the scope of our current threat model, we acknowledge their significance and leave the exploration of backdoor-resistant scenarios for future work.

\section{Conclusion}
In this paper, we proposed \ProjectName{}, a novel diffusion-based defense method to defend object detectors against both hiding and creating patch attacks. By implementing a two-stage pipeline: (i) regeneration with inpainting diffusion models, (ii) rectification through an adversarial detection algorithm, \ProjectName{} effectively neutralizes malicious patches without any prior knowledge of adversarial patches. Our evaluations on two benchmarks and real-world scenarios demonstrate that \ProjectName{} consistently outperforms SOTA defense methods across different detectors and attacks, stikes the balance between effectiveness and efficiency, and is resilient against adaptive attacks. We hope that generative model-based defenses such as \ProjectName{} will contribute to advancing the security and robustness of object detection systems.



\bibliographystyle{IEEEtran}
\bibliography{acsac}

\end{document}